\documentclass[runningheads]{llncs}

\usepackage[mobile]{eccv}

\usepackage{eccvabbrv}

\usepackage{graphicx}
\usepackage{booktabs}

\usepackage[accsupp]{axessibility}  

\usepackage[pagebackref,breaklinks,colorlinks]{hyperref}

\usepackage{orcidlink}
\usepackage{subfiles}
\usepackage{mathtools}
\usepackage{wrapfig}
\usepackage[nice]{nicefrac}

\begin{document}

\title{StyleGaussian: Instant 3D Style Transfer\\with Gaussian Splatting} 

\titlerunning{StyleGaussian}

\author{Kunhao Liu\inst{1} \and
Fangneng Zhan\inst{2} \and
Muyu Xu\inst{1} \and \\
Christian Theobalt\inst{2} \and
Ling Shao\inst{3} \and
Shijian Lu\inst{1}
}

\authorrunning{Kunhao Liu et al.}

\institute{Nanyang Technological University \and
Max Planck Institute for Informatics \and
UCAS-Terminus AI Lab, UCAS}

\maketitle

\begin{abstract}

We introduce StyleGaussian, a novel 3D style transfer technique that allows instant transfer of any image's style to a 3D scene at 10 frames per second (fps). Leveraging 3D Gaussian Splatting (3DGS), StyleGaussian achieves style transfer without compromising its real-time rendering ability and multi-view consistency. It achieves instant style transfer with three steps: embedding, transfer, and decoding. Initially, 2D VGG scene features are embedded into reconstructed 3D Gaussians. Next, the embedded features are transformed according to a reference style image. Finally, the transformed features are decoded into the stylized RGB. StyleGaussian has two novel designs. The first is an efficient feature rendering strategy that first renders low-dimensional features and then maps them into high-dimensional features while embedding VGG features. It cuts the memory consumption significantly and enables 3DGS to render the high-dimensional memory-intensive features. The second is a K-nearest-neighbor-based 3D CNN. Working as the decoder for the stylized features, it eliminates the 2D CNN operations that compromise strict multi-view consistency. Extensive experiments show that StyleGaussian achieves instant 3D stylization with superior stylization quality while preserving real-time rendering and strict multi-view consistency. Project page: \url{https://kunhao-liu.github.io/StyleGaussian/}

\end{abstract}

\section{Introduction}

\begin{figure}[t]
    \centering
    \includegraphics[width=\textwidth]{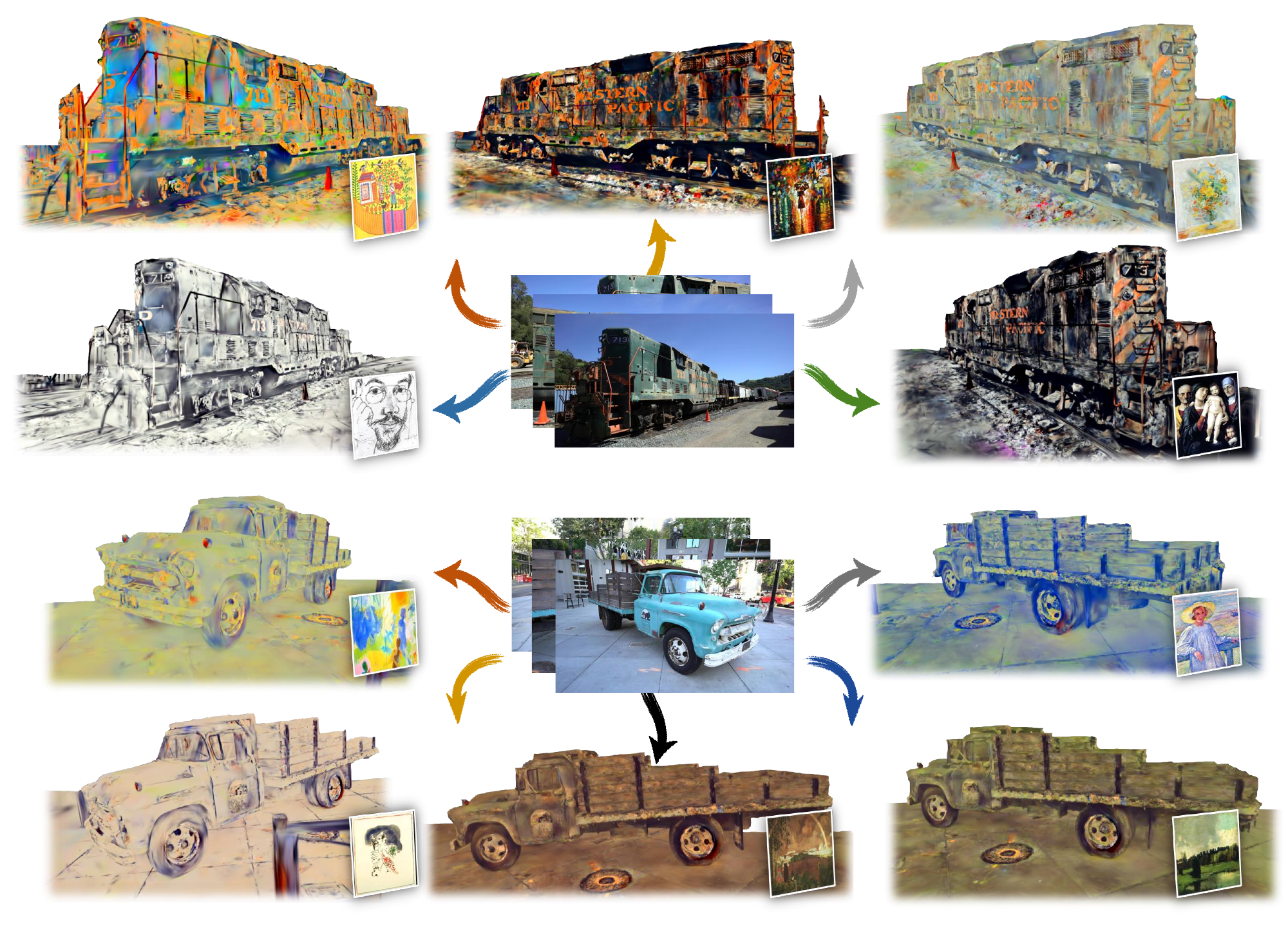}
    \caption{
    We introduce StyleGaussian, a novel 3D style transfer pipeline that enables instant style transfer while ensuring strict multi-view consistency. We mask the background for aesthetics, complete stylized novel views are shown in the experiments part.
    }
    \label{fig:teaser}
\end{figure}

The remarkable advancements in radiance fields \cite{mildenhall2021nerf} for 3D reconstruction have unveiled a new dimension for immersive exploration of the 3D world. A radiance field, mapping arbitrary 3D coordinates to color and density values, can be implemented via implicit MLPs \cite{mildenhall2021nerf}, explicit voxels \cite{sun2022direct}, or a combination of both \cite{chen2022tensorf, muller2022instant}. However, due to the inherent complexity of these implementation functions, editing a radiance field's style and appearance is not as straightforward as editing meshes or point clouds.
Under the guidance of images \cite{liu2023stylerf, zhang2022arf}, text \cite{wang2022clip}, or other forms of user input \cite{yuan2022nerf}, several studies tackle radiance field editing via learning but they necessitate test-time optimization which leads to a time-intensive editing process \cite{zhang2022arf, wang2022clip}. Alternatively, several studies introduce feed-forward networks to speed up the editing but the speed-up is limited \cite{chiang2022stylizing} and they also potentially violate multi-view consistency \cite{liu2023stylerf}.

Given the cruciality of instant interaction and feedback for a seamless 3D user experience, it's essential that the 3D style transfer can be achieved instantaneously, without sacrificing real-time rendering and multi-view consistency. 
Recently, 3D Gaussian Splatting (3DGS) \cite{kerbl20233d} has attracted increasing attention thanks to its swift rendering capabilities, making it possible to achieve instant 3D style transfer.
State-of-the-art 3D style transfer techniques involve three typical processes \cite{liu2023stylerf}: 1) feature embedding that embeds image VGG \cite{simonyan2014very} features to the reconstructed 3D geometry; 2) style transfer that transforms the embedded features based on a style image; and 3) RGB decoding that decodes the transformed features to RGB. 
Though 3DGS efficiently renders the RGB (three-dimensional) images, rendering high-dimensional features and embedding them into the reconstructed 3D geometry is both computationally and memory intensive \cite{qin2023langsplat, zhou2023feature}. In addition, most existing 3D style transfer techniques adopt 2D CNNs for decoding the stylized features to RGB, but 2D CNNs tend to impair multi-view consistency and further the quality of 3D style transfer \cite{liu2023stylerf, huang2021learning, gu2021stylenerf}.

We design StyleGaussian, a 3D style transfer pipeline that achieves instant 3D style transfer while preserving strict multi-view consistency. StyleGaussian consists of three steps: embedding, transfer, and decoding. First, we embed image VGG features \cite{simonyan2014very} into reconstructed 3D Gaussians for stylization, necessitating rendering the embedded features and comparing them with the ground truth in learning. To address the challenge of the computationally and memory intensive process of rendering high-dimensional VGG features with 3DGS, we develop an efficient feature rendering strategy that first renders low-dimensional features and then maps the rendered low-dimensional features to high-dimensional features. The strategy reformulates the mapping of 2D features as the mapping in 3D Gaussians, thereby enabling obtaining the high-dimensional VGG features for each Gaussian effectively and efficiently. 

With the acquired VGG features for each Gaussian, we can achieve stylization straightly by aligning their channel-wise mean and variance with those of the style image, akin to AdaIN \cite{huang2017arbitrary}. We then decode the stylized features back to RGB for 3DGS rendering. However, most existing work decodes the stylized features with a 2D CNN decoder which tends to compromise multi-view consistency \cite{liu2023stylerf, huang2021learning, mu20223d, gu2021stylenerf}. To this end, we design a 3D CNN that employs the K-nearest neighbors (KNN) of each Gaussian as a sliding window and formulates convolution operations to match the sliding window. As the 3D CNN works directly in the 3D space, it preserves strict multi-view consistency and leaves the rendering process of 3D Gaussians unaffected. Thanks to the above two designs, StyleGaussian allows zero-shot 3D style transfer without requiring any optimization for new style images.

The contributions of this work can be summarized in three major aspects. \textit{First}, we introduce StyleGaussian, a novel 3D style transfer pipeline that enables instant style transfer ($\sim$10 fps) while preserving real-time rendering and strict multi-view consistency. \textit{Second}, we design an efficient feature rendering strategy that enables rendering high-dimensional features while learning to embed them into reconstructed 3D Gaussians. \textit{Third}, we design a KNN-based 3D CNN that adeptly decodes the stylized features of 3D Gaussians to RGB without impairing multi-view consistency. Thorough experiments over multiple benchmarks and applications validate the superiority of our designed StyleGaussian.

\section{Related Work}
\noindent
\textbf{Radiance Fields.} 
Radiance fields \cite{mildenhall2021nerf} have recently made significant strides in advancing 3D scene representation. These fields are functions that assign radiance (color) and density values to arbitrary 3D coordinates. The color of a pixel is rendered by aggregating the radiances of 3D points through volume rendering \cite{max1995optical}. Radiance fields find extensive applications across various domains of vision and graphics, especially in view synthesis \cite{barron2021mip, barron2022mip, pumarola2021d}, generative models \cite{schwarz2020graf, niemeyer2021giraffe, chan2022efficient, gu2021stylenerf}, and surface reconstruction \cite{wang2021neus, wang2023neus2}. They can be implemented through multiple approaches, such as MLPs \cite{mildenhall2021nerf, barron2021mip, barron2022mip}, decomposed tensors \cite{chen2022tensorf, chan2022efficient, fridovich2023k}, hash tables \cite{muller2022instant}, and voxels \cite{sun2022direct, fridovich2022plenoxels}, with numerous studies aiming to enhance their quality \cite{barron2021mip, barron2022mip} or rendering and reconstruction speed \cite{chen2022tensorf, muller2022instant, fridovich2022plenoxels, reiser2021kilonerf, garbin2021fastnerf}.
Among these advancements, 3D Gaussian Splatting (3DGS) \cite{kerbl20233d} stands out for its rapid reconstruction capabilities, real-time rendering performance, and excellent reconstruction results. It models radiance fields using a multitude of explicitly parameterized 3D Gaussians. The cornerstone of its ability to render in real-time is its reliance on rasterization over ray tracing to render images. Building on the strengths of 3DGS, our work utilizes it to facilitate an immersive 3D editing experience.

\medskip
\noindent
\textbf{3D Appearance Editing.}
Editing the appearance of traditional 3D representations like meshes or point clouds is generally straightforward, as meshes are associated with UV maps and points correspond to pixels in images. However, editing radiance fields is challenging due to their implicit encoding within the parameters of neural networks or tensors. Consequently, previous studies have resorted to learning-based methods for editing radiance fields \cite{bao2023sine, liu2022nerf, liu2021editing, noguchi2021neural, wang2022clip, wang2023nerf, xu2022deforming, chen2023gaussianeditor, srinivasan2021nerv, liu2023stylerf, yuan2022nerf, zhuang2023dreameditor, haque2023instruct, chiang2022stylizing}, guided by images \cite{liu2023stylerf, bao2023sine, chiang2022stylizing, zhang2022arf}, text\cite{wang2022clip, wang2023nerf, haque2023instruct, chen2023gaussianeditor, zhuang2023dreameditor}, or other forms of user input \cite{liu2021editing, xu2022deforming, chen2023gaussianeditor}, encompassing modifications such as deformation \cite{yuan2022nerf, xu2022deforming, noguchi2021neural}, appearance changes \cite{liu2023stylerf, wang2022clip, wang2023nerf, chen2023gaussianeditor, haque2023instruct, zhuang2023dreameditor}, removal \cite{chen2023gaussianeditor}, relighting \cite{srinivasan2021nerv}, and inpainting \cite{liu2022nerf, mirzaei2023spin}.
Yet, most of these approaches rely on a test-time optimization strategy \cite{zhang2022arf, haque2023instruct, zhuang2023dreameditor, chen2023gaussianeditor, mirzaei2023spin, wang2022clip, wang2023nerf}, necessitating a time-intensive optimization process for each edit. Alternatively, some methods facilitate the editing of 3D scenes in a feed-forward manner \cite{liu2023stylerf, chiang2022stylizing}. However, the editing speed of these approaches is still far from an interactive speed.
In contrast, our method can edit the appearance of the scene instantly.

\medskip
\noindent
\textbf{Neural Style Transfer.} 
Neural style transfer aims to render a new image that merges the content structure of one image with the style patterns of another. Prior research indicates that the second-order statistics of VGG features \cite{simonyan2014very} encapsulate the style information of 2D images \cite{gatys2016image}. Initially, this field relied on optimization methods to align the style image's VGG features \cite{gatys2016image}, but subsequent approaches introduced feed-forward networks to approximate this optimization process \cite{johnson2016perceptual, li2019learning, huang2017arbitrary}, significantly enhancing the speed of style transfer. More recent efforts have extended style transfer to the 3D domain by attempting to stylize point clouds or meshes \cite{huang2021learning, mu20223d}. However, these methods often lag in rendering capabilities when compared to radiance fields, prompting further research into the stylization of radiance fields \cite{huang2022stylizednerf, chiang2022stylizing, zhang2022arf, nguyen2022snerf, fan2022unified, liu2023stylerf, wang2023nerf}. Works such as \cite{zhang2022arf, nguyen2022snerf, wang2023nerf} have managed to achieve radiance field style transfer through optimization, offering visually impressive stylizations but at the cost of time-consuming optimization for each new style and limited generalizability to unseen styles. Alternatives such as HyperNet \cite{chiang2022stylizing} embed style information into MLP parameters for arbitrary style transfer but face slow rendering and poor detail in style patterns. StyleRF \cite{liu2023stylerf} introduces zero-shot radiance field style transfer but uses a 2D CNN decoder, impairing multi-view consistency. Our method, however, allows for instant transfer and real-time rendering while maintaining strict multi-view consistency.

\section{Preliminary: 3D Gaussian Splatting}
3D Gaussian Splatting \cite{kerbl20233d} represents a 3D scene as a set of 3D Gaussian primitives $ \mathbb{G} = \{ \boldsymbol{g}_p=( \boldsymbol{\mu}_p, \boldsymbol{\Sigma}_p, \sigma_p, \boldsymbol{c}_p) \}_p^P $, where each 3D Gaussian $\boldsymbol{g}_p$ is parameterized by a mean $\boldsymbol{\mu}_p \in \mathbb{R}^3$ specifying its center, a covariance $\boldsymbol{\Sigma}_p \in \mathbb{R}^{3 \times 3}$ specifying its shape and size, an opacity $\sigma_p \in \mathbb{R}_+$ and a color $\boldsymbol{c}_p \in \mathbb{R}^3$ (we omit view-dependency for simplicity). 
3D Gaussians can be effectively rendered by computing the color $\boldsymbol{C}$ of a pixel by blending $N$ ordered Gaussians overlapping the pixel:
\begin{equation}
\label{eq:volume_render}
    \boldsymbol{C} = \sum_{i \in N} \boldsymbol{c}_i \alpha_i \prod_{j=1}^{i-1} (1-\alpha_j),
\end{equation}
where $\alpha_i$ is given by evaluating a Gaussian with its covariance $\boldsymbol{\Sigma}_i$ \cite{yifan2019differentiable} multiplied with its opacity $\sigma_i$. Gaussian Splatting provides a fast differentiable renderer that can render the 3D Gaussians in real-time. Compared to the sampling-based approaches used to render radiance fields \cite{mildenhall2021nerf, barron2021mip}, this approach is significantly more efficient in terms of time and memory. Similar to NeRF \cite{mildenhall2021nerf}, 3D Gaussians can be reconstructed on a per-scene basis, supervised with multiple posed images.

\section{StyleGaussian}

\begin{figure}[t]
    \centering
    \includegraphics[width=1\linewidth]{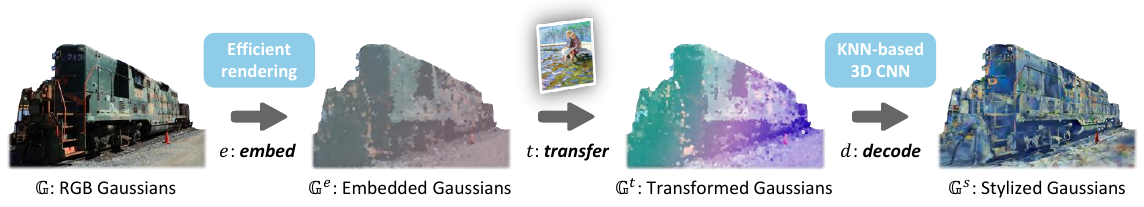}
    \caption{\textbf{Overview of StyleGaussian.} Given a reconstructed 3D Gaussians $\mathbb{G}$, we first embed the VGG features to the 3D Gaussians ($e$). Then, given a style image, we transform the features of the embedded Gaussians $\mathbb{G}^e$ to obtain $\mathbb{G}^t$, where the features are infused with the style information ($t$). Lastly, we decode the transformed features of $\mathbb{G}^t$ into RGB to produce the final stylized 3D Gaussians $\mathbb{G}^s$ ($d$). We design an efficient feature rendering strategy in $e$ that enables rendering high-dimensional VGG features while learning to embed them into $\mathbb{G}$. We also develop a KNN-based 3D CNN as the decoder in $d$. }
    \label{fig:overview}
\end{figure}

We present StyleGaussian, a 3D style transfer method that can transfer the style of any image to reconstructed 3D Gaussians instantly. Given the reconstructed 3D Gaussians $\mathbb{G} = \{\boldsymbol{g}_p\}_p^P$ of a scene, and a style image $I^s$, our goal is to stylize the color of each Gaussian such that the stylized 3D Gaussians $\mathbb{G}^s = \{\boldsymbol{g}_p^s\}_p^P$ capture the style of $I^s$ while keeping their original content structure. For each stylized Gaussian $\boldsymbol{g}_p^s$, we fix its geometry and only transform the color: $\boldsymbol{g}_p^s : \boldsymbol{c}_p \rightarrow \boldsymbol{c}^s_p$. The stylized 3D Gaussians $\mathbb{G}^s$ have the same rendering process as the original 3D Gaussians $\mathbb{G}$, thus maintaining real-time rendering and strict multi-view consistency. 

We denote the style transfer process as $f: f(\mathbb{G}, I^s) = \mathbb{G}^s$. Straightforwardly, $f$ can be implemented with a neural network that handles sets of data or point clouds \cite{qi2017pointnet, qi2017pointnet++}. However, 3D Gaussians are a scene-specific model and usually contain considerable primitives ($\geq 10^5$), making it sub-optimal to implement $f$ as a single large neural network. Inspired by the zero-shot style transfer literature \cite{huang2017arbitrary, liu2023stylerf}, we decompose $f$ into three steps, as shown in \cref{fig:overview}:  \textbf{1) embedding} $e$, which embeds the 2D VGG image features into $\mathbb{G}$: 
\begin{equation}
    e(\mathbb{G}) = \mathbb{G}^e = \{ \boldsymbol{g}^e_p=( \boldsymbol{\mu}_p, \boldsymbol{\Sigma}_p, \sigma_p, \boldsymbol{f}_p) \}_p^P ,
\end{equation}
where each embedded Gaussian $\boldsymbol{g}^e_p$ has its VGG feature $\boldsymbol{f}_p$;  \textbf{2) transfer} $t$, which transforms the features of the embedded 3D Gaussians $\mathbb{G}^e$ according to the input style image $I^s$: 
\begin{equation}
    t(\mathbb{G}^e, I^s) = \mathbb{G}^{t} = \{ \boldsymbol{g}^{t}_p=( \boldsymbol{\mu}_p, \boldsymbol{\Sigma}_p, \sigma_p, \boldsymbol{f}^t_p) \}_p^P ,
\end{equation}
where the transformed feature of each Gaussian $\boldsymbol{f}^t_p$ has been infused with the style information of $I^s$;  \textbf{3) decoding} $d$, which decodes the transformed features of 3D Gaussians back to RGB:
\begin{equation}
    d(\mathbb{G}^{t}) = \mathbb{G}^s = \{ \boldsymbol{g}_p^s = ( \boldsymbol{\mu}_p, \boldsymbol{\Sigma}_p, \sigma_p, \boldsymbol{c}^s_p) \}_p^P.
\end{equation}
Note that the geometry properties $\boldsymbol{\mu}_p, \boldsymbol{\Sigma}_p, \sigma_p $ of each Gaussian $\boldsymbol{g}_p$ are fixed during the whole style transfer process, only the feature of each Gaussian changes: $\boldsymbol{c}_p \xrightarrow[]{e} \boldsymbol{f}_p \xrightarrow[]{t} \boldsymbol{f}^t_p \xrightarrow[]{d} \boldsymbol{c}^s_p$. In summary, the whole style transfer process can be described as $\{ \boldsymbol{c}_p \xrightarrow[]{f} \boldsymbol{c}^s_p ; I^s \}_p^P, f = e \circ t \circ d$.  We will elaborate on the feature embedding process $e$ in \cref{sec:encoder}, the style transfer process $t$ in \cref{sec:transferrer}, and the RGB decoding process $d$ in \cref{sec:decoder}.

\subsection{Feature Embedding}
\label{sec:encoder}

The VGG features of a 2D image are obtained by extracting features from the intermediate layers of a pre-trained VGG network \cite{simonyan2014very}. To embed the VGG features to a radiance field, following \cite{liu2023stylerf, kobayashi2022decomposing}, we distill the features from the multi-view 2D VGG feature maps. Given that Gaussian Splatting is a scene-specific model, we can embed the VGG features for 3D Gaussians after reconstruction and keep them fixed during the stylization process.  We describe the feature embedding process $e$ below.

Specifically, given the reconstructed 3D Gaussians of a scene, we assign each Gaussian $\boldsymbol{g}_p$ a learnable feature parameter $\boldsymbol{f}_p \in \mathbb{R}^D $, and render the feature of each pixel $\boldsymbol{F} \in \mathbb{R}^D$ in the same way as rendering the RGB (\cref{eq:volume_render})\footnote{We use subscript $p$ for $\boldsymbol{f}_p$ when we refer to the $p$th Gaussian in the whole set of 3D Gaussians, and we use subscript $i$ for $\boldsymbol{f}_i$ when we refer to the $i$th Gaussian along a ray during volume rendering. They are the same if referring to the feature of the same Gaussian.}:
\begin{equation}
\label{eq:feature_render}
    \boldsymbol{F} = \sum_{i \in N} \boldsymbol{f}_i w_i, \quad
    w_i = \alpha_i \prod_{j=1}^{i-1} (1-\alpha_j),
\end{equation}
where $w_i$ is the blending weight of each Gaussian. We then optimize $ \{ \boldsymbol{f}_p \}_p^P$ using the L1 loss objective between the rendered features and the ground truth VGG features:
$\{ \boldsymbol{f}_p \}_p^P = \underset{\{ \boldsymbol{f}_p \}_p^P}{\text{argmin}} \sum |\boldsymbol{F}- \boldsymbol{F}_{gt}|$.

\begin{wrapfigure}{R}{0.5\textwidth}
    \centering
    \includegraphics[width=\linewidth]{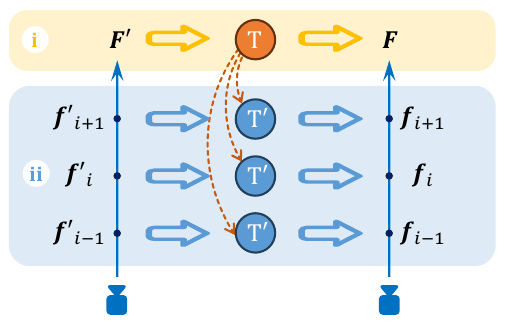} 
    \caption{\textbf{Efficient feature rendering.} We first render the low-dimensional features $\boldsymbol{F}'$ and then map them to high-dimensional features $\boldsymbol{F}$. Part $\mbox{\textbf{i}}$ corresponds to \cref{eq:affine}, where $\mbox{T}$ denotes the affine transformation applied to $\boldsymbol{F}'$. Part $\mbox{\textbf{ii}}$ corresponds to \cref{eq:derivation}, where $\mbox{T}'$ denotes the affine transformation applied to $\boldsymbol{f}'_i$. $\mbox{T}$ can be reformulated to $\mbox{T}'$, enabling the derivation of the high-dimensional feature $\boldsymbol{f}_i$ from $\boldsymbol{f}'_i$ for each Gaussian. }
    \label{fig:encoder}

\end{wrapfigure}

However, as VGG features have a large number of channels ($D=256$ in \verb+ReLU3_1+ layer and $D=512$ in \verb+ReLU4_1+ layer), rendering such high-dimensional features directly through Gaussian Splatting is both computationally and memory intensive \cite{qin2023langsplat, zhou2023feature}. In fact, even when $D=256$, the renderer cannot be compiled because it exceeds the shared memory limits of the GPU\footnote{We use NVIDIA RTX A5000 with 24G memory in our experiments, whose performance is beyond the average of most consumer GPUs.}. Thus, we propose an efficient feature rendering strategy, rendering low-dimensional features $\boldsymbol{F}' \in \mathbb{R}^{D'}, D'=32$ before mapping them to high-dimensional features $\boldsymbol{F}$. The learnable feature parameter of each Gaussian is now $\boldsymbol{f}'_p \in \mathbb{R}^{D'}$, and the low-dimensional feature of each pixel can be rendered as $\boldsymbol{F}' = \sum_{i \in N} \boldsymbol{f}'_i w_i$.

The key problem here is how to map $\boldsymbol{F}'$ to $\boldsymbol{F}$ while obtaining the high-dimensional VGG feature $\boldsymbol{f}_p$ of each Gaussian. To address this, we design the mapping process as an affine transformation $\mbox{T}()$:
\begin{equation}
\label{eq:affine}
    \boldsymbol{F} = \mbox{T}(\boldsymbol{F}') = \boldsymbol{A} \boldsymbol{F}' + \sum_{i \in N} \boldsymbol{b}w_i,
\end{equation}
where $\boldsymbol{A} \in \mathbb{R}^{D \times D'} $ is the learnable linear transformation, $\boldsymbol{b} \in \mathbb{R}^D$ is the learnable bias, and $w_i$ is the blending weight in \cref{eq:feature_render}. The optimization objective is now: $ \boldsymbol{A}, \boldsymbol{b}, \{ \boldsymbol{f}'_p \}_p^P = \underset{\boldsymbol{A}, \boldsymbol{b}, \{ \boldsymbol{f}'_p \}_p^P}{\text{argmin}} \sum |\boldsymbol{F}- \boldsymbol{F}_{gt}| $. Since our goal is to obtain $\boldsymbol{f}_p$ and we now only possess the optimized low-dimensional features $\boldsymbol{f}'_p$, we proceed to derive the relationship between $\boldsymbol{f}'_p$ and $\boldsymbol{f}_p$. Note that \cref{eq:affine} can be written as:
\begin{equation}
\label{eq:derivation}
    \boldsymbol{F} = \boldsymbol{A} \sum_{i \in N} \boldsymbol{f}'_i w_i + \sum_{i \in N} \boldsymbol{b}w_i
    = \sum_{i \in N} ( \boldsymbol{A} \boldsymbol{f}'_i + \boldsymbol{b}) w_i = \sum_{i \in N} \mbox{T}'( \boldsymbol{f}'_i ) w_i,
\end{equation}
where $\mbox{T}'()$ donotes the affine transformation applied to $\boldsymbol{f}'_i$. With \cref{eq:feature_render} and \cref{eq:derivation}, it is obvious that $\boldsymbol{f}_p = \mbox{T}'( \boldsymbol{f}'_p ) =  \boldsymbol{A} \boldsymbol{f}'_p + \boldsymbol{b}$. The visualization of the feature rendering process is in \cref{fig:encoder}. We now obtain the VGG features $\{ \boldsymbol{f}_p \}_p^P$ of the 3D Gaussians. The VGG features of 3D Gaussians are pre-optimized and fixed during stylization, thus the embedding process does not affect the speed of style transfer.

\subsection{Style Transfer}
\label{sec:transferrer}
After obtaining the 3D Gaussians $\mathbb{G}^e$ embedded with VGG features, now we need to transform the features according to an arbitrary input style image $I^s$. Since our goal is real-time style transfer, we opt for a parameter-free and efficient zero-shot style transfer algorithm, AdaIN \cite{huang2017arbitrary}, to implement the style transfer process $t$. AdaIN performs style transfer by aligning the channel-wise mean and variance of content features with those of the input style features. Specifically, the transformed feature $\boldsymbol{f}_p^t \in \mathbb{R}^D$ of a Gaussian is given by:
\begin{equation}
    \boldsymbol{f}_p^t = \sigma(\boldsymbol{F}^s) \left( \frac{\boldsymbol{f}_p - \mu( \{ \boldsymbol{f}_p \}_p^P  )}{\sigma( \{ \boldsymbol{f}_p \}_p^P)}  \right) + \mu(\boldsymbol{F}^s),
\end{equation}
where $\mu() \in \mathbb{R}^D , \sigma() \in \mathbb{R}^D$ denote the channel-wise mean and standard deviation, $\boldsymbol{F}^s$ denotes the VGG features of the input style image $I^s$. We thus obtain the transformed 3D Gaussians $\mathbb{G}^t$ where each Gaussian's feature $\boldsymbol{f}_p^t$ has been infused with the style information of $I^s$.

\subsection{RGB Decoding}
\label{sec:decoder}

\begin{wrapfigure}{L}{0.45\textwidth}
\includegraphics[width=\linewidth]{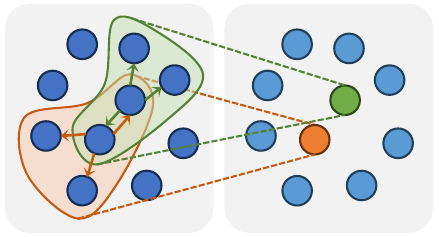} 
\caption{\textbf{Illustation of the KNN-based convolution.} We treat each Gaussian’s KNN as the sliding window. Left represents the Gaussians in one layer, right represents the Gaussians in the next layer. }
\label{fig:3dconv}
\end{wrapfigure}

Following the acquisition of the transformed 3D Gaussians $\mathbb{G}^t$, we proceed to decode the features into RGB to obtain the final stylized 3D Gaussians $\mathbb{G}^s$ in the decoding process $d$. Unlike previous methods that first render the feature maps and then decode them into stylized RGB images \cite{liu2023stylerf, huang2021learning, mu20223d}, we decode the features into RGB directly in 3D, and subsequently render the stylized RGB images. This approach maintains the rendering process of 3D Gaussians unchanged, ensuring real-time rendering and strict multi-view consistency. Although it is possible to use a per-point MLP to decode the features, previous research has indicated that the decoder must possess a large receptive field to accurately decode the stylized RGB \cite{liu2023stylerf, huang2021learning, mu20223d}. This necessity precisely explains why previous studies have preferred a 2D CNN over an MLP for decoding the features.

We design a simple and effective 3D CNN as the decoder for decoding the transformed features to the stylized RGB. The decoder processes each Gaussian's K-nearest neighbors (KNN) within a sliding window and is composed of multiple layers, as shown in \cref{fig:3dconv}. The output channel of the final layer is set to 3, meaning the final output from the decoder corresponds to the stylized RGB values. For each Gaussian $\boldsymbol{g}_p^t$, we denote its KNN as $ \{  \boldsymbol{g}_k^t  \}_k^K $, noting that the center Gaussian itself is also included in its KNN. Therefore, the input features for the first layer of the sliding window centered at $\boldsymbol{g}_p^t$ are $ \{  \boldsymbol{f}_k^t \}_k^K $. For each layer, we refer to the input features of a sliding window in this layer as $\{  \boldsymbol{f}_k^{in} \in \mathbb{R}^{D_{in}} \}_k^K$. The output of the sliding window, assigned to the center Gaussian $ \boldsymbol{g}_p^t $, is denoted as $ \boldsymbol{f}_p^{out} \in \mathbb{R}^{D_{out}} $. The parameters of the convolution operations in a layer include weights and bias terms, with the weights denoted as $\{  \boldsymbol{W}_k \in \mathbb{R}^{D_{out} \times D_{in}}   \}_k^K$ and the bias as $ \boldsymbol{B} \in \mathbb{R}^{D_{out}} $. Consequently, the convolution operation in the sliding window centered at $\boldsymbol{g}_p^t$ can be expressed as:
\begin{equation}
\label{eq:3D_conv}
    \boldsymbol{f}_{p}^{out} = \phi \left( \sum_{k=1}^K \boldsymbol{W}_k \boldsymbol{f}_k^{in} + \boldsymbol{B} \right),
\end{equation}
where $\phi$ denotes the Sigmoid function. This convolution can be efficiently implemented through matrix multiplication. We stack the features of all $P$ Gaussians' KNN to form the input feature tensor $\boldsymbol{F}_{in} \in \mathbb{R}^{P \times K \times D_{in}}$, which is subsequently reshaped to a matrix $\tilde{\boldsymbol{F}}_{in} \in \mathbb{R}^{P \times (K \times D_{in}) }$. Similarly, we also stack all $K$ weights to form $\boldsymbol{W} \in \mathbb{R}^{D_{out} \times D_{in} \times K} $, which is then reshaped to another matrix $\tilde{\boldsymbol{W}} \in \mathbb{R}^{D_{out} \times ( D_{in} \times K)} $. Hence, we can then perform the convolution of \cref{eq:3D_conv} as:
\begin{equation}
\label{eq:conv_mul}
    \boldsymbol{F}_{out} = \phi \left( \tilde{\boldsymbol{F}}_{in} \tilde{\boldsymbol{W}}^T + \dot{\boldsymbol{B}} \right) \in \mathbb{R}^{P \times D_{out}},
\end{equation}
where $T$ denotes transpose, $\dot{\boldsymbol{B}}$ indicates that $\boldsymbol{B}$ is broadcasted to $\mathbb{R}^{P \times D_{out}}$. The operations described in \cref{eq:3D_conv} and \cref{eq:conv_mul} are equivalent, but \cref{eq:conv_mul} is easier to implement to be executed parallelly on GPU. Consequently, we obtain the stylized 3D Gaussians $\mathbb{G}^s$, which are capable of being rendered in real-time while maintaining strict multi-view consistency. For any given style reference image, the style transfer process is performed only once. After acquiring $\mathbb{G}^s$, we are able to render stylized novel views without the need to repeat the style transfer process. We use the same training loss as \cite{huang2017arbitrary} in the stylization training: $ \mathcal{L} = \mathcal{L}_{c} + \lambda \mathcal{L}_{s} $, where the content loss $\mathcal{L}_{c}$ is the MSE of the feature maps, the style loss $\mathcal{L}_{s}$ is the MSE of the channel-wise feature mean and standard deviation, and $\lambda$ balances the content preservation and the stylization effect.

\section{Experiments}

\begin{figure}
    \centering
    \includegraphics[width=\linewidth]{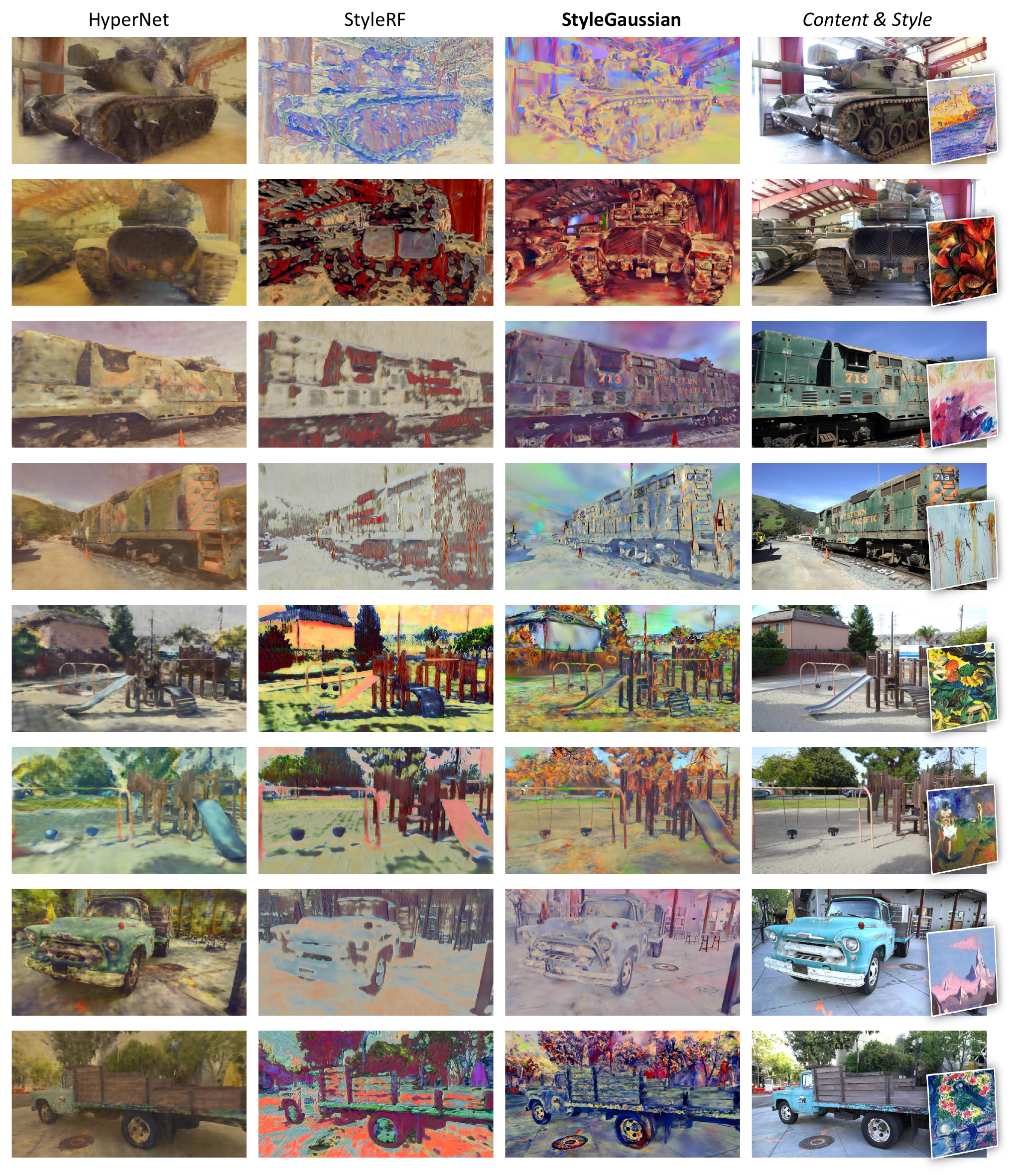}
    \caption{\textbf{Qualitative results.} Comparison of StyleGaussian with two zero-shot radiance field style transfer methods: HyperNet \cite{chiang2022stylizing} and StyleRF \cite{liu2023stylerf}. StyleGaussain demonstrates superior style transfer quality with better style alignment with the style reference images and better content preservation.}
    \label{fig:comparsion}
\end{figure}

\noindent
\textbf{Datasets and Baselines.} 
We evaluate StyleGaussian on two real-world scene datasets: the Tanks and Temples dataset \cite{knapitsch2017tanks} and the Mip-NeRF 360 dataset \cite{barron2022mip}. Both datasets contain real-world unbounded captures with complex geometries and intricate details. We use WikiArt \cite{WikiArt} as the training style images dataset. We also experiment with a diverse set of style images to test our method's ability to handle a wide range of style exemplars.
We compare our method with two state-of-the-art zero-shot radiance field style transfer methods. HyperNet \cite{chiang2022stylizing} uses implicit MLPs (i.e. NeRF$++$\cite{zhang2020nerf++}) to reconstruct a radiance field for a scene and then updates the weights of the radiance prediction branch using a hypernetwork that takes a style image as input. On the other hand, StyleRF \cite{liu2023stylerf} utilizes TensoRF \cite{chen2022tensorf} to represent a radiance field and embeds VGG features into the reconstructed radiance field. Unlike our method, StyleRF first renders the 2D feature maps and then transforms and decodes the features. We do not compare with optimization-based style transfer methods, as their transfer time is significantly longer than zero-shot methods (taking more than 10 minutes per transfer) \cite{zhang2022arf, nguyen2022snerf}. We also do not compare with mesh- or point-cloud-based style transfer methods, as previous work has shown that they struggle to reconstruct the correct scene geometry \cite{liu2023stylerf, huang2021learning}.

\begin{table}[t]

\caption{\textbf{Quantitative results.} 
We evaluate the performance of StyleGaussian against the state-of-the-art in terms of consistency, using LPIPS (\(\downarrow\))  and RMSE (\(\downarrow\)). Additionally, we assess its performance regarding transfer time and rendering time (excluding IO time).}
\centering

\begin{tabular}{@{\hspace{0.7em}}c@{\hspace{0.7em}}|@{\hspace{0.7em}}c@{\hspace{0.7em}}c@{\hspace{0.7em}}c@{\hspace{0.7em}}c@{\hspace{0.7em}}|@{\hspace{0.7em}}c@{\hspace{0.7em}}c@{\hspace{0.7em}}}
\toprule

Methods & 
\multicolumn{2}{c}{\begin{tabular}{@{}c@{}}Short-range\\Consistency\end{tabular}} & 
\multicolumn{2}{c|@{\hspace{0.7em}}}{\begin{tabular}{@{}c@{}}Long-range\\Consistency\end{tabular}} & 
\begin{tabular}{@{}c@{}}Transfer\\Time\end{tabular} & 
\begin{tabular}{@{}c@{}}Rendering\\Time\end{tabular} \\

\midrule

{}  & \textit{LPIPS} & \textit{RMSE}  & \textit{LPIPS} & \textit{RMSE} & \textit{Seconds} & \textit{Seconds} \\

HyperNet \cite{chiang2022stylizing} &  0.036  &  0.043 &  0.076  &  0.078 & 1.908 & 32.373 \\
StyleRF \cite{liu2023stylerf} &  0.050  & 0.045  &  0.123 &  0.098 & 3.403 & 10.084 \\
\textbf{StyleGaussian} &  \textbf{0.026}  & \textbf{0.031} & \textbf{0.072} &  \textbf{0.073} & \textbf{0.105} & \textbf{0.005} \\
\bottomrule
\end{tabular}

\label{tab:comparison}
\end{table}

\subsection{Qualitative Results}
We show the qualitative results in \cref{fig:comparsion}. StyleGaussain demonstrates superior style transfer quality with better style alignment with the style reference images and better content preservation. HyperNet fails to capture the correct styles, showing monotonic color and over-smooth appearance. This is largely due to the hypernetwork's limited ability to capture the correct styles of arbitrary images. 
StyleRF performs better in style transfer compared to HyperNet. However, it still struggles to capture the style in rows 3, 7, and 8, showing different colors with the reference style images. Furthermore, it produces unwanted artifacts and striped textures in the stylized novel views, which are likely due to the use of 2D CNN.
In contrast, StyleGaussian produces stylized novel views capturing the correct styles of the reference style images with fewer artifacts and better content preservation. The superior stylization is largely attributed to our design that directly decodes the stylized RGB in 3D.

\subsection{Quantitative Results}

As pointed out in previous work \cite{zhang2022arf, liu2023stylerf}, there still does not exist a standard quantitative metric for evaluating 3D style transfer quality. Thus, we perform quantitative comparisons on multi-view consistency, transfer, and rendering time, as shown in \cref{tab:comparison}. For multi-view consistency, following previous work \cite{chiang2022stylizing, fan2022unified, huang2021learning, mu20223d, liu2023stylerf}, we warp one view to another according to the optical flow \cite{teed2020raft} using softmax splatting \cite{niklaus2020softmax}, and then compute the masked RMSE score and LPIPS score \cite{zhang2018unreasonable} to measure stylization consistency. StyleGaussian outperforms StyleRF by a large margin, as StyleGaussian uses a 3D CNN to decode the RGB values, maintaining strict multi-view consistency, whereas StyleRF uses a 2D CNN for decoding, which impairs multi-view consistency. StyleGaussian also outperforms HyperNet, since HyperNet retains a view-dependency effect that compromises consistency in the style transfer task.

Furthermore, StyleGaussian significantly outperforms others in both transfer time and rendering speed. Specifically, StyleGaussian achieves a transfer time speed-up of 18 times compared to HyperNet and 32 times compared to StyleRF. This rapid transfer capability is primarily due to our method's design, which applies style transfer to the entire 3D representation at once and leverages a 3D CNN for decoding the transformed features to RGB values. 
Most notably, StyleGaussian maintains the real-time rendering capability of 3DGS, as it requires the style transfer process to be performed only once for a given style reference image. Afterward, it allows for the rendering of stylized novel views without the need to repeat the style transfer process. In contrast, HyperNet and StyleRF intertwine the rendering and style transfer processes. They need to conduct style transfer for each new view rendering, further slowing down the already sluggish rendering speeds of NeRF$++$ and TensoRF. Quantitatively, StyleGaussian significantly enhances rendering speed, achieving speed-ups of 6000 times and 2000 times compared to HyperNet and TensoRF, respectively.

\subsection{Ablations}

\begin{figure}[t]
    \centering
    \includegraphics[width=\linewidth]{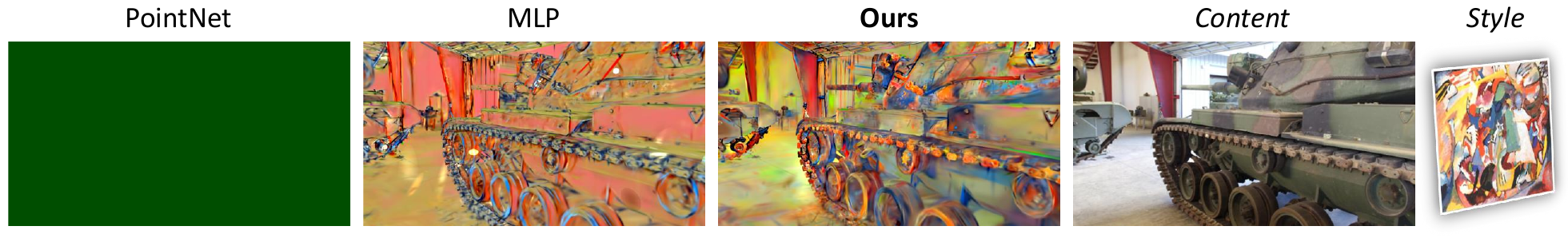}
    \caption{\textbf{Decoder design.} We validate the effectiveness of the KNN-based 3D CNN by comparing it with alternatives like MLP or PointNet.}
    \label{fig:ablation}
\end{figure}

\begin{figure}[t]
    \centering
    \includegraphics[width=\linewidth]{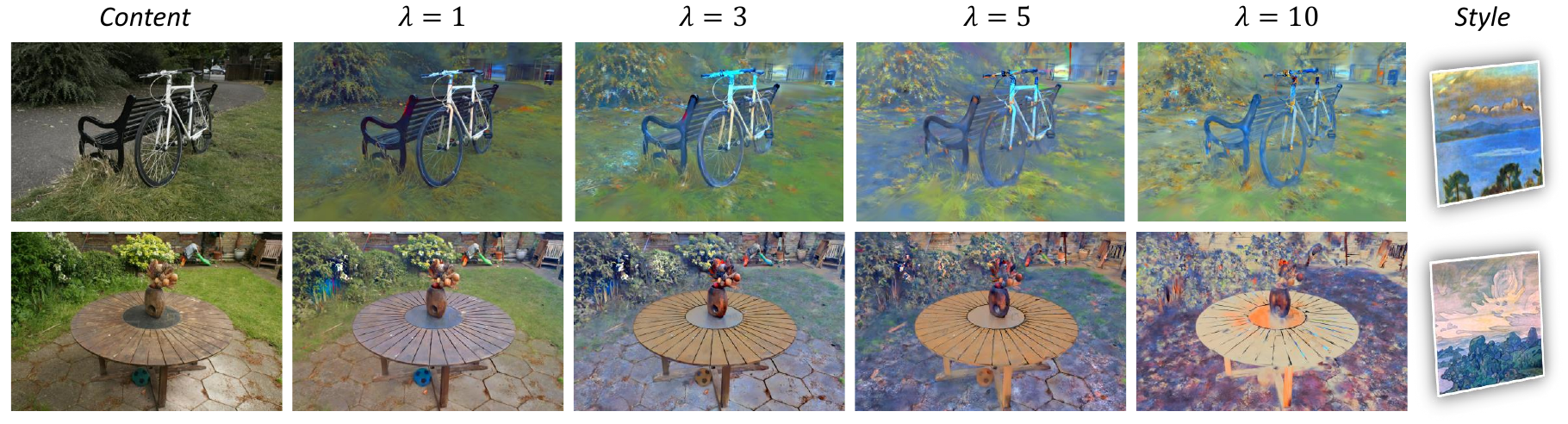}
    \caption{\textbf{Content-style balance.} We can balance content preservation and stylization effect by adjusting $\lambda$.}
    \label{fig:content_interpolation}
\end{figure}

\noindent
\textbf{KNN-based 3D CNN.} 
We validate the effectiveness of the KNN-based 3D CNN by comparing it with alternatives like MLP or PointNet \cite{qi2017pointnet}, as demonstrated in Fig. \ref{fig:ablation}. Despite our best efforts, PointNet still fails to produce meaningful color variations, resulting in each Gaussian displaying the same color. This issue likely stems from the implementation of the symmetry functions, which cause a group of points to produce identical outputs.  While MLP can generate colors that somewhat reflect the intended stylization, its per-point receptive field size limits its ability to transfer more general style patterns like strokes and texture. These limitations highlight the superior capability of the KNN-based 3D CNN in these aspects.

\medskip
\noindent
\textbf{Content-style balance.} 
We investigate the impact of $\lambda$ for the content-style balance, as shown in \cref{fig:content_interpolation}.  By adjusting $\lambda$ during training, we can opt to retain more content details or to emphasize style patterns and colors. In our experiments, we find that $\lambda = 10$ yields visually most pleasant results.

\subsection{Applications}

\begin{figure}[t]
    \centering
    \includegraphics[width=\linewidth]{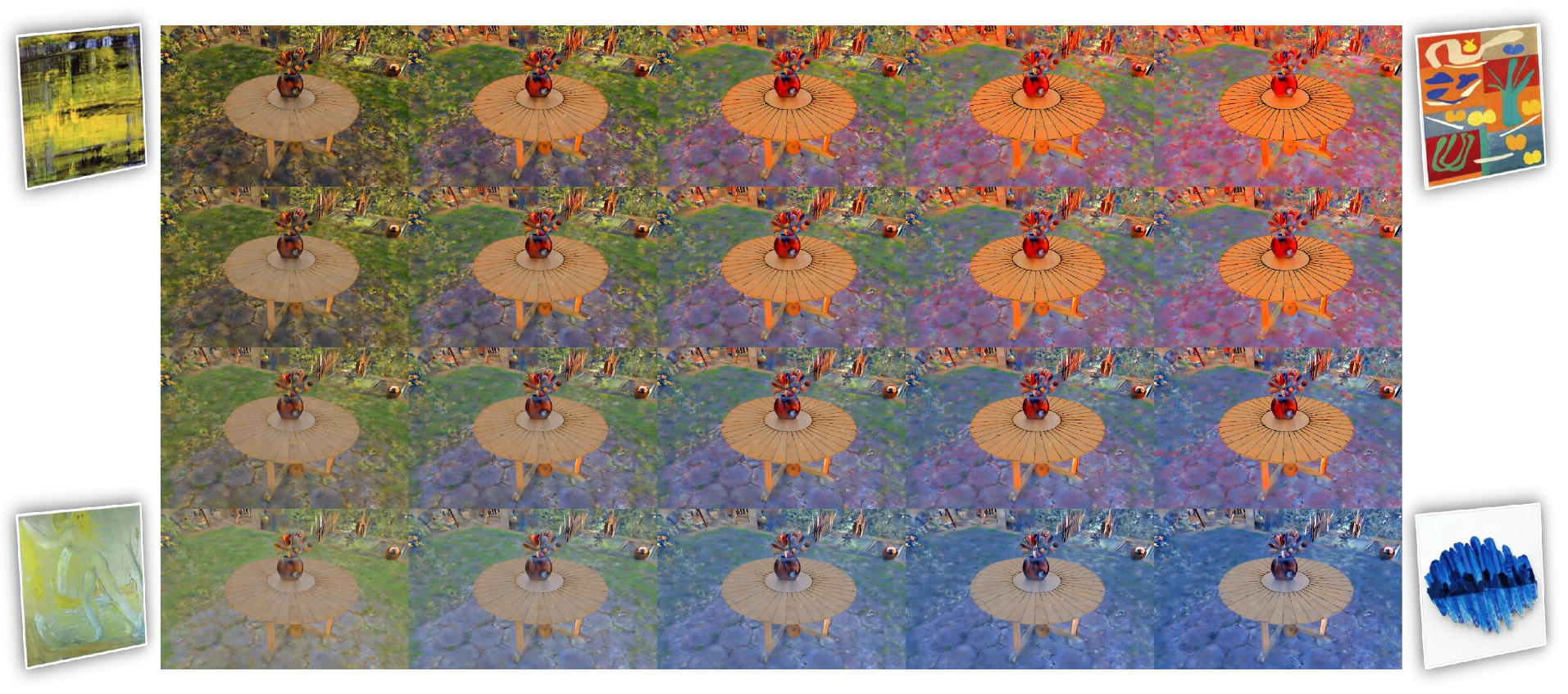}
    \caption{\textbf{Style interpolation.} StyleGaussian can smoothly interpolate between different styles.}
    \label{fig:style_interpolation}
\end{figure}

\begin{figure}[t]
    \centering
    \includegraphics[width=\linewidth]{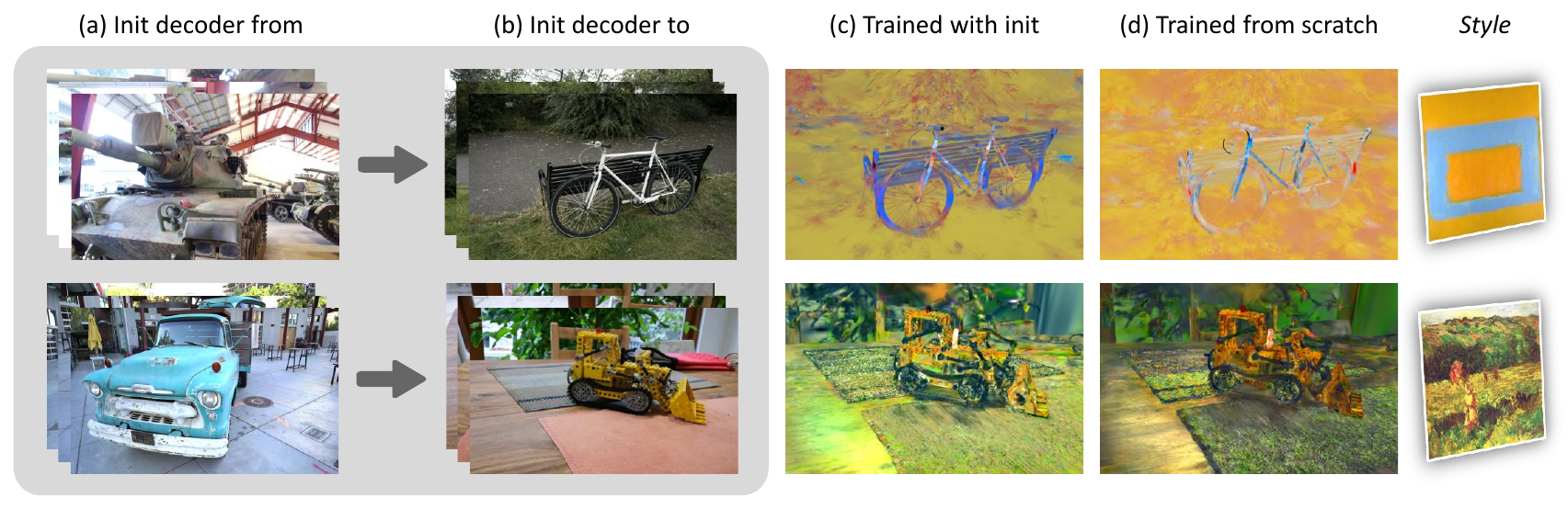}
    \caption{\textbf{Decoder initialization.} We initialize the decoder for the scene (b) using the trained decoder from the scene (a). The stylization results of the decoder trained with initialization (c) can resemble those from the decoder trained from scratch (d) but with less than $\nicefrac{1}{3}$ of the training time.}
    \label{fig:new_scene}
\end{figure}

\noindent
\textbf{Style interpolation.} 
StyleGaussian enables smooth interpolation between different styles at inference time, as demonstrated in \cref{fig:style_interpolation}. This interpolation is achieved by blending the transformed features $\mathbb{G}^t$ corresponding to different styles and subsequently decoding the interpolated features into RGB using the decoder. The interpolated stylized Gaussian can still be rendered in real-time with strict multi-view consistency.

\medskip
\noindent
\textbf{Decoder initialization.} 
One constraint common to all zero-shot radiance field style transfer methods, including StyleGaussian, is the requirement for time-consuming pre-scene training. To mitigate this issue, we propose initializing the decoder of a new scene with a trained decoder from another scene, as depicted in \cref{fig:new_scene}. We can then fine-tune the decoder instead of training it from scratch. This initialization process allows us to achieve similar stylization results with less than $\nicefrac{1}{3}$ of the training time.

\section{Limitations}

Our method has two major limitations. First, StyleGaussian only modifies the color of 3D Gaussians but leaves the geometry unchanged. Thus, distortions or some style patterns cannot be faithfully transferred. As most zero-shot radiance field style transfer methods currently focus only on stylizing the color while fixing the geometry, exploring zero-shot geometry style transfer for radiance fields presents an interesting direction. Second, while the number of parameters in StyleGaussian's style transfer module is independent of the number of Gaussians, both memory and computation costs increase with the number of Gaussians. Thus we limit the number of Gaussians to $3 \times 10^5$ for a 24G memory capacity. Future research could explore how to achieve zero-shot style transfer without storing the high-dimensional VGG features, which could save considerable memory.

\section{Conclusion}

In this paper, we introduce StyleGaussian, a novel 3D style transfer method capable of instantly transferring the style of any image to reconstructed 3D Gaussians of a scene. StyleGaussian achieves style transfer without compromising the real-time rendering capabilities and multi-view consistency of 3D Gaussian Splatting. The style transfer process consists of three steps: embedding, transfer, and decoding, which embeds VGG features, stylizes these features, and decodes the features to RGB, respectively. We introduce an efficient feature rendering strategy to address the challenge of rendering and embedding high-dimensional features. Additionally, we develop a KNN-based 3D CNN as the decoder, which possesses a large receptive field and maintains strict multi-view consistency. StyleGaussian has the potential to be integrated into various applications, including augmented reality, virtual reality, video games, and film production, by enabling the seamless integration of artistic styles into 3D environments.

%
%
\bibliographystyle{splncs04}
\bibliography{egbib}

\clearpage

\appendix

\noindent{\LARGE \bfseries Appendix}
\vspace{1em}

This document provides supplementary materials for \textit{StyleGaussian: Instant 3D Style Transfer with Gaussian Splatting} including implementation details and additional results. We strongly encourage readers to watch the \textbf{attached video }(no speed-up; all style transfer and style interpolation happen on the fly), which demonstrates the instant style transfer ability and stylization quality of StyleGaussian.

\section{Implementation Details}

\noindent \textbf{Feature embedding training.}
In the feature embedding process \(e\), we utilize features extracted from the \verb+ReLU3_1+ layer of the VGG network, indicating that \(D=256\). Since the extracted VGG feature maps differ in resolution from the RGB images, we render the feature maps to match the resolution of the VGG feature maps directly, avoiding interpolation to conserve memory. The model is trained for 30,000 iterations during the feature embedding process. We set the learning rate for the learnable features, \(\{ \boldsymbol{f}'_p \}_k^K \), to 0.01, and the learning rate for the parameters of the affine transformation, \(\boldsymbol{A}\) and \(\boldsymbol{b}\), to 0.001.

\medskip
\noindent \textbf{Style transfer training.}
As we utilize a parameter-free style transfer algorithm, AdaIN, for efficiency, the decoder is the only component that requires optimization during the style transfer training process. We define $K=8$ within the KNN-based 3D CNN and construct a sequence of 5 convolutional layers with output channels set to 256, 128, 64, 32, and 3, respectively. Given that the geometry of the 3D Gaussians remains constant during stylization, it is only necessary to compute the KNN for each Gaussian once. The model is trained for 100,000 iterations in the style transfer training process. We set the learning rate for the decoder to 0.001. When the decoder is initialized from a previously trained decoder of another scene, the training duration for the decoder is reduced to 30,000 iterations. For the computation of both content loss and style loss, we employ the \verb+ReLU1_1+, \verb+ReLU2_1+, \verb+ReLU3_1+, and \verb+ReLU4_1+ layers of the VGG network.

\section{Additional Results}
\begin{figure}
    \centering
    \includegraphics[width=\linewidth]{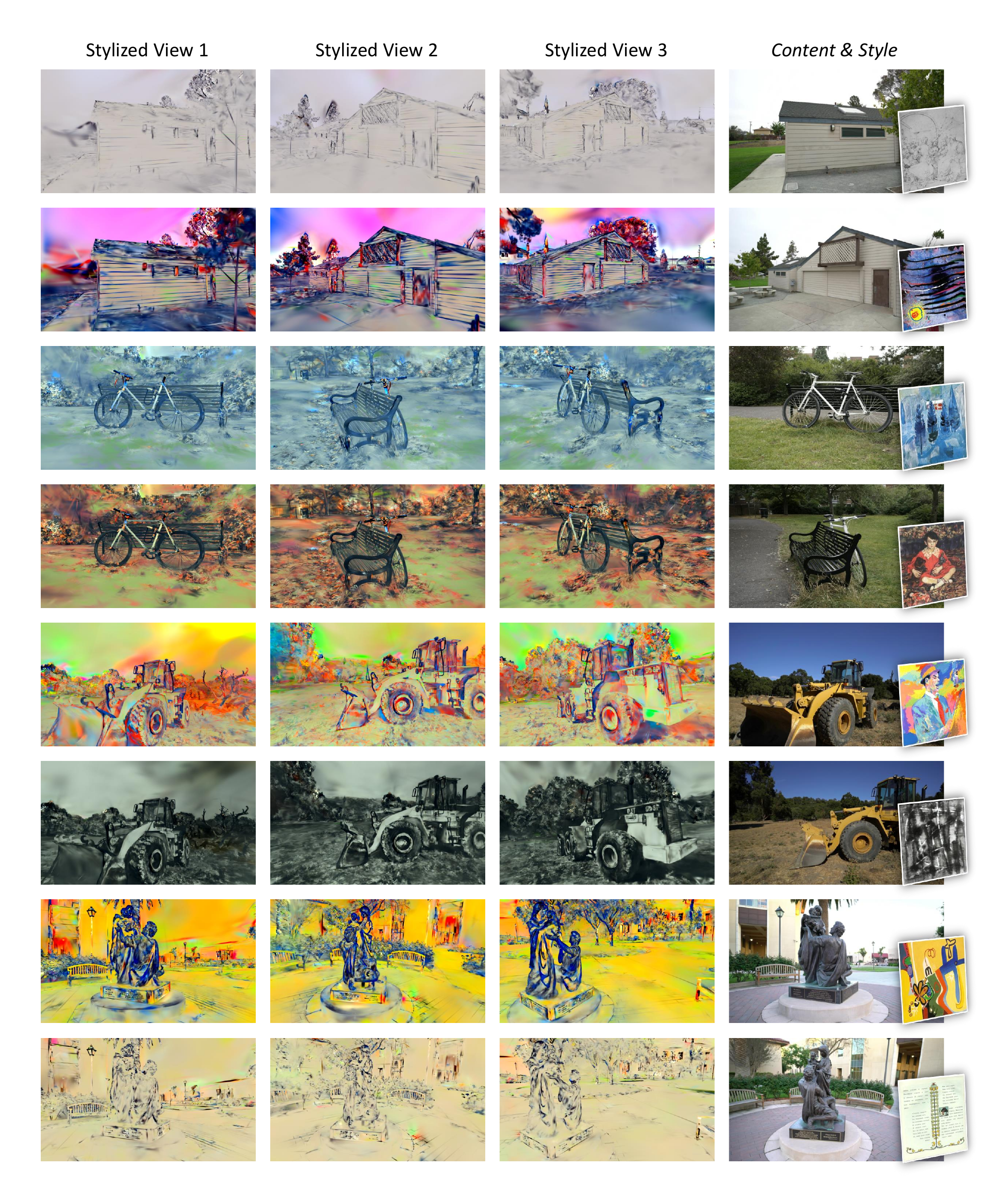}
    \caption{\textbf{Additional stylization results in more scenes.} All styles are transferred instantly in a zero-shot manner.}
    \label{fig:results1}
\end{figure}

\begin{figure}
    \centering
    \includegraphics[width=\linewidth]{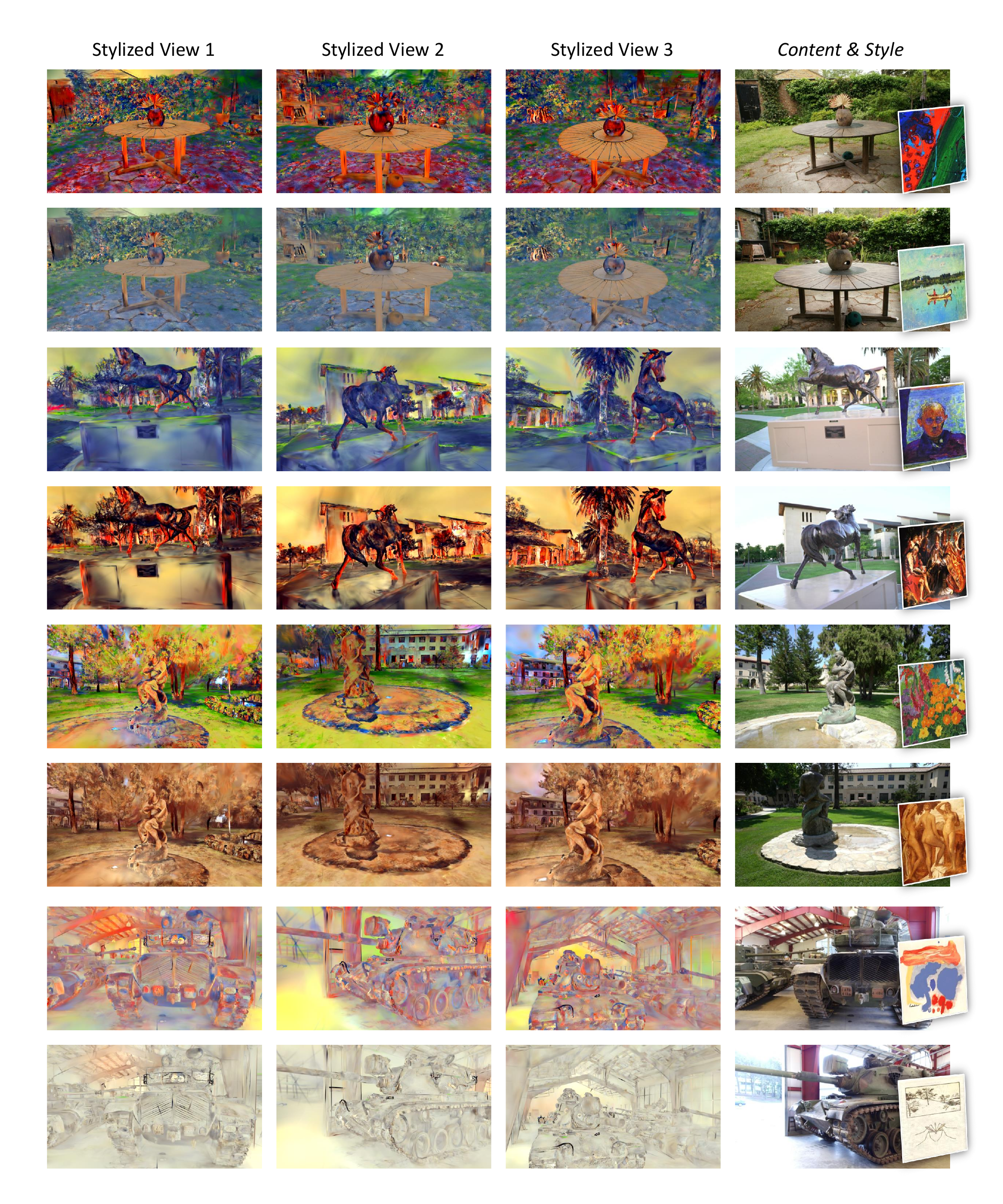}
    \caption{\textbf{Additional stylization results in more scenes.} All styles are transferred instantly in a zero-shot manner.}
    \label{fig:results2}
\end{figure}

We provide multiview images showcasing additional stylization results across various scenes in \cref{fig:results1} and \cref{fig:results2}.

\end{document}